\documentclass[conference]{IEEEtran}
\IEEEoverridecommandlockouts

\usepackage{cite}
\usepackage{amsmath,amssymb,amsfonts}
\usepackage{graphicx}
\usepackage{textcomp}
\usepackage{xcolor}
\usepackage{algorithm}
\usepackage{array}
\usepackage{colortbl}
\usepackage{multirow}
\usepackage{algpseudocode}
\usepackage{algorithmicx}
\usepackage{adjustbox}
\usepackage{multirow}
\usepackage{booktabs}
\usepackage{hhline}
\usepackage{subcaption}

\def\BibTeX{{\rm B\kern-.05em{\sc i\kern-.025em b}\kern-.08em
    T\kern-.1667em\lower.7ex\hbox{E}\kern-.125emX}}
\begin{document}

\title{DA-MoE: Towards Dynamic Expert Allocation for Mixture-of-Experts Models\\
{\footnotesize \textsuperscript{}}
}

\author{
    \IEEEauthorblockN{Maryam Akhavan Aghdam, Hongpeng Jin, Yanzhao Wu}
    \IEEEauthorblockA{\textit{Knight Foundation School of Computing and Information Sciences} \\
    \textit{Florida International University}\\
    Miami, FL, United States \\
    makha003@fiu.edu, hjin008@fiu.edu, yawu@fiu.edu}
}

\maketitle

\begin{abstract}
Transformer-based Mixture-of-Experts (MoE) models have been driving several recent technological advancements in Natural Language Processing (NLP). These MoE models adopt a router mechanism to determine which experts to activate for routing input tokens. However, existing router mechanisms allocate a fixed number of experts to each token, which neglects the varying importance of different input tokens. In this study, we propose a novel dynamic router mechanism that Dynamically Allocates a variable number of experts for Mixture-of-Experts (DA-MoE) models based on an effective token importance measure.
First, we show that the Transformer attention mechanism provides a natural and effective way of calculating token importance.
Second, we propose a dynamic router mechanism that effectively decides the optimal number of experts (K) and allocates the top-K experts for each input token.
Third, comprehensive experiments on several benchmark datasets demonstrate that our DA-MoE approach consistently outperforms the state-of-the-art Transformer-based MoE model on the popular GLUE benchmark.

\end{abstract}


\section{Introduction}

\begin{figure}
    \centering
    \includegraphics[width=1\linewidth]{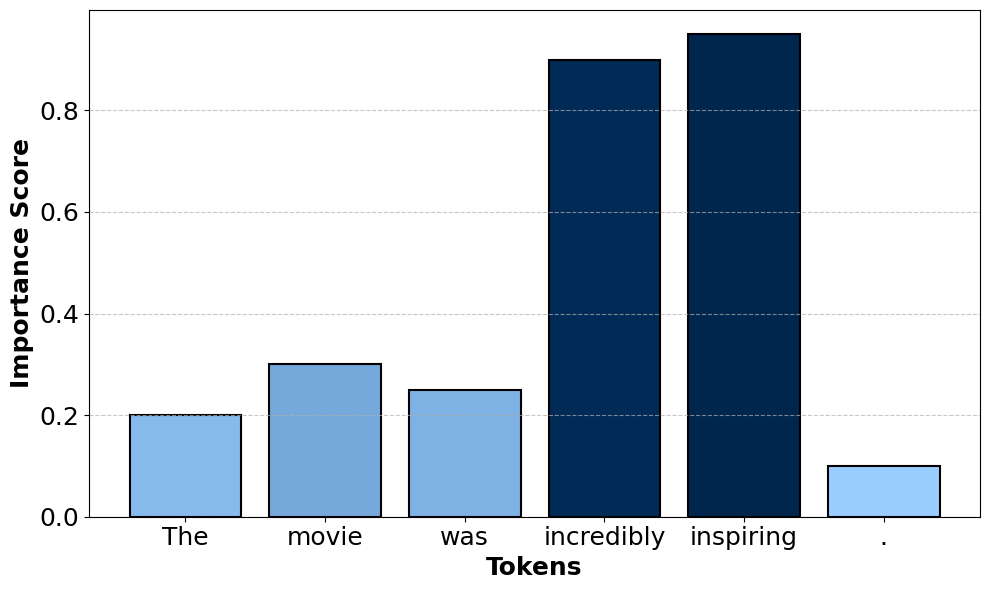}
    \caption{An example input sentence (``The movie was incredibly inspiring.") for performing sentiment analysis}
    \label{fig:token-importance-ex}
\end{figure}

Natural Language Processing (NLP) has achieved remarkable success over the last several years, fueled by big data and emerging large-scale deep neural networks, particularly Transformer-based neural networks~\cite{b1,b2,b5,b27,b28,b29,b30,b31,b32,b33,b34,b35}.
Several recent studies~\cite{b22,b23,b24,b25,b26} have shown that larger neural networks tend to be more sample-efficient and deliver better generalization with enhanced predictive performance. 
One prominent approach to scaling Transformer models is the Mixture-of-Experts (MoE), which utilizes multiple subnetworks (experts) and sparsity to efficiently scale Transformer architectures to trillions of model parameters~\cite{b3,b4,b5}. 
MoE models adopt a router to determine which experts to activate to process each token. 
Existing MoE methods suffer from the allocation of a fixed number of expert networks for each input token, which may not dynamically adapt to the varying importance of different input tokens.
For example, Figure~\ref{fig:token-importance-ex} illustrates an example input sentence (``The movie was incredibly inspiring.") for performing sentiment analysis. In this case, certain tokens (i.e., ``incredibly" and ``inspiring") carry substantial semantic weight and play a critical role in accurate model predictions, requiring the activation of more experts to capture their latent semantics. On the other hand, less informative tokens (e.g., ``The" and ``as") can be allocated to fewer experts, potentially enhancing overall model efficiency.
This observation motivates us to explore a novel research question in this study, i.e., how to determine token importance and leverage the token importance to dynamically allocate experts for Mixture-of-Experts models.

We first briefly introduce Transformers, MoE models, and their integration. Transformers represent a widely used deep neural network architecture powered by self-attention, which enables the model to assess the importance of different words or tokens in an input sequence, capture long-range dependencies, and focus on key parts of the input to effectively capture contextual information and relationships among input tokens~\cite{b1,b2,b4,b5,b45,b46,b47,b48}. This design makes them highly effective for processing sequence data, particularly in natural language process tasks~\cite{b1,b2,b4,b5,b48}. Hence, Transformers have been adopted by numerous state-of-the-art language models, such as BERT~\cite{b2}, GPT-3~\cite{b26}, ChatGPT~\cite{b23}, and Llama~\cite{b24}. 
Mixture-of-Experts~\cite{b7,b4,b5,b49,b50,b51,b52,b57,b58,b66,b67} is a sparsely activated neural network where a router selects only a subset of experts (subnetworks) for processing each input instead of utilizing all model parameters, thereby improving computational efficiency and scalability. 
The incorporation of MoE techniques into Transformer models has achieved early success in scaling Transformers to trillions of model parameters~\cite{b4,b5}. However, existing methods simply allocate a fixed number of experts for input tokens without accounting for their varying importance, leaving it still an open challenge to design effective MoE models that consider token importance to further enhance the predictive performance and efficiency of large-scale Transformers.
The attention layer calculates attention weights by determining the relevance of each token in a sequence using query, key, and value vectors, allowing the model to focus on important tokens based on context.
This attention mechanism in Transformer architectures potentially provides a natural way to capture token importance, which can guide the MoE router in determining the specific number of experts and which experts to activate for achieving dynamic expert allocation.

Bear this objective in mind, we propose a new Mixture-of-Experts model, \textbf{D}ynamic \textbf{A}ttention-based \textbf{M}ixture-\textbf{o}f-\textbf{E}xperts (DA-MoE). DA-MoE leverages the attention mechanism in Transformers to assess the importance of each input token and redesign the router to dynamically activate the top-K experts for each input token, where K is determined dynamically based on the token importance. This design enables the dynamic expert allocations for Mixture-of-Experts models and significantly enhances model predictive performance and overall efficiency. Concretely, we made three novel contributions in this paper. 
\begin{enumerate}
     \item We analyze and identify the key limitations of conventional Mixture-of-Experts models, that is, the allocation of a fixed number of experts to each input token. This approach fails to account for the varying importance of input tokens, resulting in inefficient resource usage and sub-optimal predictive performance. 
     \item We propose a novel dynamic router mechanism for Mixture-of-Experts models, which can effectively leverage the Transformer attention mechanism to evaluate token importance and dynamically allocate experts for input tokens based on their importance measures.  
     \item We conduct comprehensive experiments on popular datasets to evaluate the proposed DA-MoE model under both training and fine-tuning scenarios. The experimental results demonstrate that DA-MoE can effectively scale to a large number of experts and significantly outperform the baseline Transformer model on the representative General Language Understanding Evaluation (GLUE) benchmark~\cite{b8}.
 \end{enumerate} 

 The rest of this paper is organized as follows. Section~\ref{section:related-work} reviews the related studies in the literature. Section~\ref{section:problem-statement} describes the research problems that we aim to address in this study. Section~\ref{section:methods} provides a detailed description of our proposed DA-MoE approach. Section~\ref{section:experimental-analysis} presents our experimental analysis on representative benchmarks. Finally, Section~\ref{section:conclusion} concludes this paper with our findings and outlines potential directions for future research.

\section{Related Work} \label{section:related-work}
Transformer models, empowered by the self-attention mechanism, have revolutionized sequence data processing and attracted numerous efforts to optimize their efficiency, scalability, and adaptability to meet the increasing demands of various real-world applications~\cite{b1,b2,b4,b5,b43,b44,b62,b63,b64,b65}. We briefly summarize three broad categories of techniques to optimize Transformer architectures. 
The \textit{first} category of efforts aims to enhance the Transformer attention mechanisms to reduce the inherent high computation costs due to the quadratic complexity with respect to the input sequence length. Models, like Linformer~\cite{b41}, Reformer~\cite{b33}, and Performer~\cite{b42}, optimize Transformer attention mechanisms to reduce the computational costs while retaining the core benefits of self-attention.
The \textit{second} category of efforts focuses on reducing the number of parameters in Transformer models without compromising model predictive performance~\cite{b28,b31,b59}. For example, ALBERT~\cite{b28} introduces parameter sharing across layers, substantially reducing the model size while maintaining competitive performance.
The \textit{third} category of efforts is represented by various Mixture-of-Experts (MoE) techniques, which have been highly effective in scaling Transformers to trillions of parameters without proportional increases in computational costs~\cite{b4,b5,b7,b53,b54,b55,b56,b60,b61}. This study contributes to the third category with a novel DA-MoE model with dynamic expert allocation mechanisms.

\begin{figure*}[h!]
    \centering
    \includegraphics[width=1\linewidth]{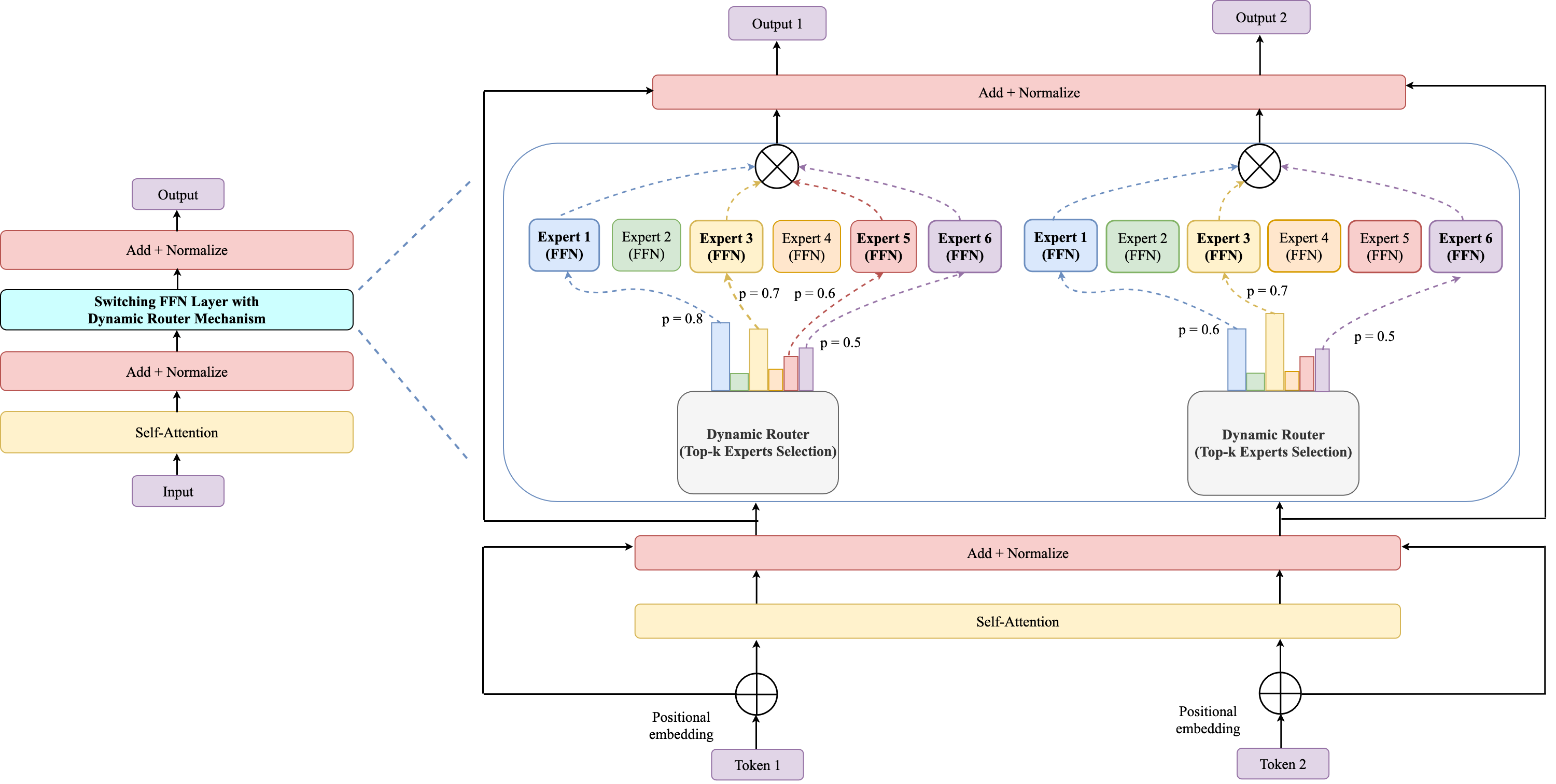}
    \caption{Illustration of a DA-MoE encoder block with a dynamic routing mechanism. DA-MoE introduces a dynamic routing mechanism, allowing each token to be assigned to the top-K experts based on token importance. For example, token one is assigned to four experts, while token two is assigned to three experts.}
    \label{fig:MoE-architecture}
\end{figure*}

The MoE models have been pushing forward several technological advancements in deep learning, especially in scaling natural language processing models~\cite{b3,b4,b5,b36,b39,b40}. The concept of MoE was first introduced by Jacobs et al. in the early 1990s to train multiple experts with specialization in different regions of the input space~\cite{b7}. At the core of MoE models is the router mechanism (gating network), which selects the appropriate expert(s) to activate for processing each input.
Several studies have proposed router mechanisms with different design principles and strategies for assigning input tokens to experts. Below, we summarize the representative router mechanisms for MoE models.
\begin{enumerate}
    \item Sparsely-Gated MoE~\cite{b3}, introduced by Shazeer et al., selects a sparse subset of experts based on the input. Even though the gating network can determine which experts to choose, the number of selected experts remains a pre-defined hyperparameter during training and inference. 
    \item GShard~\cite{b4}, developed by Lepikhin et al., scales the Sparsely-Gated MoE models to trillions of model parameters by distributing the computation across multiple computing devices. GShard demonstrates the feasibility of training large-scale MoE models and the importance of sparsity by employing a fixed router mechanism with the top-2 experts per input.
    \item Switch Transformers (ST)~\cite{b5}, proposed by Fedus et al., represents one of the milestones in MoE models, which only selects the top-1 expert per token based on the probabilities produced by the router via a softmax layer. Switch Transformer substantially improves the MoE training efficiency and scalability with state-of-the-art performance on various natural language processing tasks.
\end{enumerate}

\section{Problem Statement} \label{section:problem-statement}
Existing MoE models allocate a fixed number of experts K to each input token, regardless of its contextual relevance and importance in the task. This fixed allocation can be expressed as:
\begin{equation}
\label{eq:fixed_attention}
\mathbf{A}_{\text{fixed}}(t_i) = K
\end{equation}
where \( t_i \) represents the \( i \)-th input token and \( K \) is a constant number of experts assigned to each token.

However, tokens in an input sequence may contribute unevenly to the model prediction. Some tokens, such as named entities or task-relevant keywords, carry more semantic importance. While others, such as stop words, are less important. Under a fixed allocation scheme, both high- and low-importance tokens activate the same number of experts, resulting in inefficient expert allocation and reduced performance.

In contrast, a dynamic expert allocation mechanism assigns experts based on the contextual importance of each token, denoted as \( I(t_i) \). The number of experts for each token would now vary according to its token importance, leading to the following formulation:
\begin{equation}
\label{eq:dynamic_attention}
\mathbf{A}_{\text{dynamic}}(t_i) = f(\mathcal{I}(t_i))
\end{equation}
where \( f(I(t_i)) \) is a function that dynamically determines the number of experts based on the token importance score \( I(t_i) \), which can be computed via the Transformer attention mechanisms or other importance measures.

To address the limitations of allocating a fixed number of experts in MoE models, we propose the DA-MoE, a novel dynamic allocation MoE model leveraging an attention-based routing mechanism. Our DA-MoE model dynamically evaluates the importance of each token and adjusts the number of experts accordingly.

\begin{figure*}[t]
\centering
\makebox[\textwidth][c]{ 
\begin{minipage}{0.88\textwidth} 
\begin{algorithm}[H]
    \caption{Dynamic Expert Allocation for Mixture-of-Experts Models}
    \label{alg:dynamic-expert-allocation-MoE}
    1: \textbf{Input:} Token embeddings $X \in \mathbb{R}^d$, attention weights $A \in \mathbb{R}^{B \times H \times L \times D}$, router weights \\ \hspace*{6mm} $W_r \in \mathbb{R}^{E \times d}$, batch size ($B$), number of heads ($H$), sequence length ($L$), total number of experts \\ \hspace*{6mm} ($E$), embedding dimensions ($D$), expert capacity ($C$). \\
    2: \textbf{Output:} Aggregated output from selected experts. \\
    3: \textbf{Calculate Token Importance Scores:}\\
    4: \quad $token\_importance_k = \frac{1}{H} \sum_{j=1}^{H} \max_{d \in [1, D]} A_{i,j,k,d}$ \\
    5: \textbf{Determine the Number of Experts to Route Each Token:}\\
    6: \quad $num\_experts\_to\_route = \left\lceil token\_importance \times E \right\rceil$\\
    7: \textbf{Compute the Routing Probabilities for Each Expert:}\\
    8: \quad $router\_logits = W_r \cdot X$\\
    9: \quad $router\_probs_i = \frac{\exp(router\_logits_i)}{\sum_{j=1}^{E} \exp(router\_logits_j)}$ \text{for } $i = 1, 2, \dots, E$\\
    10: \textbf{Select Top-K Experts for Each Token:}\\
    11: \quad $(expert\_gate, expert\_index) = \text{top-K}(router\_probs, K = num\_experts\_to\_route)$\\
    12: \textbf{Generate Expert Mask:}\\
    13: \quad $expert\_mask_i = \sum_{k=1}^{num\_experts\_to\_route} \{expert\_index_k = i\}$ \text{for } $i = 1, 2, \dots, E$\\
    14: \textbf{Satisfy Expert Capacity Constraints:}\\
    15: \quad $position\_in\_expert_i = \sum_{j=1}^{i} expert\_mask_j$ \text{for } $i = 1, 2, \dots, E$\\
    16: \quad \parbox[t]{\dimexpr\linewidth-\algorithmicindent}{$expert\_mask_i = expert\_mask_i \times \{position\_in\_expert_i < C\}$\\\hspace{\algorithmicindent}\text{for } $i = 1, 2, \dots, E$}
\end{algorithm}
\end{minipage}
}
\end{figure*}

\section{DA-MoE Overview} 
\label{section:methods}

We propose \textbf{D}ynamic \textbf{A}ttention-based \textbf{M}ixture-\textbf{o}f-\textbf{E}xperts (DA-MoE), which leverages the Transformer attention mechanism to measure token importance. This design allows DA-MoE models to dynamically allocate experts for input tokens with varying importance. This section presents an overview of our DA-MoE model, including the model architecture and algorithm description.

\subsection{DA-MoE Model Architecture}
We present our DA-MoE architecture in Figure~\ref{fig:MoE-architecture}. In order to dynamically allocate experts for tokens of varying importance, we redesign the router mechanism to incorporate attention-based token importance measurements and dynamic expert allocation. Experts are allocated based on the token importance scores, with more experts (i.e., more computational resources) assigned to tokens with higher importance scores to more accurately capture the latent semantic information, while fewer experts (i.e., less computational resources) will be activated for less informative tokens.
This token importance based dynamic routing mechanism represents the key innovation of our DA-MoE models, enabling efficient allocation of computing resources for tokens with different importance and potentially improving overall scalability and efficiency. Key components of our DA-MoE approach include:

\begin{enumerate}
    \item \textbf{Attention-based Token Importance Measure:}
    The Transformer attention layer computes the scaled dot-product attention weights for each input token, which reflects how much focus each input token should receive relative to other tokens in the same input sequence~\cite{b1}. Tokens with higher attention weights are considered more influential within the input sequence, requiring the model to concentrate more on extracting latent semantics from them \cite{b1}. Hence, attention weights naturally serve as an effective way to define token importance, which allows our DA-MoE model to dynamically allocate computational resources (the number of experts) to the most critical tokens. 
    \item \textbf{Dynamic Router Mechanism:} 
    We redesign the router mechanism in MoE models to achieve dynamic expert allocation for each input token. Concretely, the token importance scores guide the router in determining the number of experts (K) to activate for each input token. The router then computes the probabilities for all available experts and routes the input token to the top-K experts ranked by these probabilities. This design ensures that more experts (i.e., more computational resources) are allocated for processing tokens deemed more important, which can effectively enhance the overall efficiency.
    \item \textbf{Other Transformer Layers:} 
    A key advantage of our proposed DA-MoE approach is that it does not require modifications to other Transformer layers, such as Multi-Head Self-Attention (MHSA)~\cite{b1}. By simply replacing the dense Feed-Forward Network (FNN) with a sparse Switching FNN integrated with our dynamic router mechanism, Transformer models can benefit from dynamic expert allocation with enhanced predictive performance and efficiency.
\end{enumerate}

\subsection{Dynamic Expert Allocation Algorithm}

We then introduce the proposed \textbf{D}ynamic Expert \textbf{A}llocation approach for \textbf{M}ixture-\textbf{o}f-\textbf{E}xperts models (DA-MoE). Algorithm~\ref{alg:dynamic-expert-allocation-MoE} presents a sketch of the key steps, including (1) calculating token importance scores based on Transformer attention weights, (2) determining the number of experts to route each input token based on token importance scores, (3) computing the routing probabilities for each expert, (3) selecting top-K experts for each input token, (4) generating the expert masks, and (5) satisfying expert capacity constraints.
This design allows our DA-MoE to dynamically allocate experts based on token importance and improve overall predictive performance and efficiency. We describe each step in detail below.

\noindent \textbf{Calculate Token Importance Scores:} Before the routing process, we leverage the attention weights from the last attention layer to calculate the token importance scores with Equation~\ref{eq:token_importance}.
\begin{equation}
\label{eq:token_importance}
token\_importance_k = \frac{1}{H} \sum_{j=1}^{H} \max_{d \in [1, D]} A_{i,j,k,d}
\end{equation}
where, for each input token $k$ in the input sequence at position $i$, we first identify the maximum attention weight it receives from each attention head $j$, which reflects its strongest connection to another token indexed by $d$ across the embedding dimensions. It offers several benefits to extract the maximum attention weight for each token in each head, such as capturing the most critical relationship with other tokens, filtering out noises, and improving interpretability. Ultimately, this leads to enhanced measurements of token importance. Then, we average these maximum attention weights across all attention heads $H$ to produce an importance score for each input token $k$ (at position $i$), which reflects the relative importance of each token in the entire sequence, with higher scores indicating tokens that play a more significant role in the model prediction process.

\noindent \textbf{Determine the Number of Experts to Route Each Token:} DA-MoE then utilizes the token importance score to determine the number of experts each token should be routed to. Specifically, we multiply the importance score of each token by the total number of all available experts ($E$) and round the result up to the nearest integer (Equation~\ref{eq:num_experts_to_route}).
\begin{equation}
\label{eq:num_experts_to_route}
num\_experts\_to\_route = \left\lceil token\_importance \times E \right\rceil
\end{equation}

\noindent \textbf{Compute the Routing Probabilities for Each Expert:} Next, we compute the routing probabilities for each expert. Following the common practice in the literature~\cite{}, we first calculate the router logits through the multiplication of the router weights $\mathrm{W_r}$ with the token embeddings $\mathrm{X}$ (Equation~\ref{eq:router_logits}). The router probabilities are then obtained by applying the softmax function to the router logits $\mathrm{E}$ (Equation 4).
\begin{equation}
\label{eq:router_logits}
router\_logits = W_r \cdot X
\end{equation}
\begin{equation}
\label{eq:router_probs}
\begin{split}
router\_probs_i &= \frac{\exp(router\_logits_i)}{\sum_{j=1}^{E} \exp(router\_logits_j)} \\
&\text{for } i = 1, 2, \dots, E
\end{split}
\end{equation}

\noindent \textbf{Select Top-K Experts for Each Token:} Once the number of experts ($num\_experts\_to\_route$) is determined for each input token (see Equation~\ref{eq:num_experts_to_route}), the router chooses the top-K experts to process the input token, where K=$num\_experts\_to\_route$. Based on the router probabilities computed for each expert (see Equation~\ref{eq:router_probs}, we can rank all experts and select the top-K experts following Equation~\ref{eq:expert_gate_index}.
\begin{multline}
\label{eq:expert_gate_index}
\hspace*{0.2\linewidth}(expert\_gate, expert\_index) = \\
\texttt{top-K}(router\_probs, K = num\_experts\_to\_route)
\end{multline}

\noindent \textbf{Generate Expert Mask:}
The next step is to generate the expert mask, specifying which experts are selected for each input token (see Equation~\ref{eq:expert_mask}).
\begin{equation}
\label{eq:expert_mask}
\begin{aligned}
expert\_mask_i &= \! \! \! \! \! \! \! \! \! \! \! \! \! \! \! \! \!\sum_{k=1}^{num\_experts\_to\_route} \! \! \! \! \! \! \! \! \! \! \! \! \! \! \! \! \! \{expert\_index_k = i\} \\
&\hspace{0.5em}\text{for } i = 1, 2, \dots, E
\end{aligned}
\end{equation}

\noindent \textbf{Satisfy Expert Capacity Constraints:}
The final step ensures the expert assignments satisfy the expert capacity constraints. In order to achieve balanced workloads across different experts, we follow~\cite{b7} to calculate the expert capacity ($C$), perform a cumulative sum operation to track the position of each token within its assigned experts ($position\_in\_expert_i$ in Equation~\ref{eq:position_in_expert}), and ensure the expert assignments will not exceed the expert capacity (see Equation~\ref{eq:expert_mask_update}). The resulting expert mask, adjusted for capacity, effectively controls the assignment of tokens across different experts, preventing overload while allowing more experts to be activated for tokens of higher importance.
\begin{equation}
\label{eq:position_in_expert}
\begin{aligned}
position\_in\_expert_i &= \!\sum_{j=1}^{i} expert\_mask_j \\
&\text{for } i = 1, 2, \dots, E
\end{aligned}
\end{equation}
\begin{multline}
\label{eq:expert_mask_update}
\hspace*{0.35\linewidth} expert\_mask_i = \\
expert\_mask_i \times \{position\_in\_expert_i < C\} \\
\hfill \text{for } i = 1, 2, \dots, E \hfill
\end{multline}

\section{Experimental Analysis} \label{section:experimental-analysis} 

In our prototype implementation, we integrate the proposed DA-MoE into one of the state-of-the-art Mxiture-of-Experts Transformer models, Switch Transformer (ST)~\cite{b5}, which serves as the baseline method for comparison.
We perform several sets of experiments to evaluate the proposed DA-MoE approach, including (1) \textbf{pre-training evaluation} on the WikiText-103 dataset~\cite{b9}, (2) \textbf{fine-tuning evaluation} with pre-trained models on the C4 (Colossal Clean Crawled Corpus) dataset~\cite{b20}, and (3) \textbf{token importance analysis} across various natural language processing tasks, such as sentiment analysis and paraphrasing. We leverage the General Language Understanding Evaluation (GLUE) benchmark~\cite{b8} to evaluate the performance of all models across various natural language processing tasks after fine-tuning, following the benchmark specifications. All experiments are conducted with 8 NVIDIA A100 GPUs.

\begin{table*}[h!]
\centering
\caption{Fine-tuning results on GLUE subtasks (pre-trained on WikiText-103 dataset). The best result for each task is highlighted in bold. M, B, and E represent million, billion, and experts.}
\label{table:pre-training-results-WikiText-103}
\begin{adjustbox}{max width=\textwidth}
\begin{tabular}{|c|c|c|c|c|c|c|c|c|}
\hline
\multirow{2}{*}{\textbf{GLUE Subtasks}} & \multirow{2}{*}{\textbf{Metrics}} & \multicolumn{2}{c|}{\textbf{600M/8E}} & \multicolumn{2}{c|}{\textbf{1B/16E}} & \multicolumn{2}{c|}{\textbf{2B/32E}} \\ \cline{3-8} 
 &  & \textbf{Baseline} & \textbf{Our Approach} & \textbf{Baseline} & \textbf{Our Approach} & \textbf{Baseline} & \textbf{Our Approach} \\ \hline\hline
\textbf{CoLA} & Accuracy & 62.89 & \textbf{64.71} & 65.29 & \textbf{68.03} & 66.71 & \textbf{69.22} \\ \hline
\textbf{SST-2} & Accuracy & 77.35 & \textbf{77.99} & 78.12 & \textbf{78.4} & 79.19 & \textbf{79.81} \\ \hline
\textbf{MRPC} & F1 & \textbf{74.01} & 68.05 & \textbf{75.87} & 68.98 & \textbf{76.29} & 69.84 \\ \hline
\textbf{QQP} & F1 & 67.7 & \textbf{76.14} & 67.83 & \textbf{76.31} & 68.93 & \textbf{77.51} \\ \hline
\textbf{MNLI} & Accuracy & 61.5 & \textbf{63.8} & 62.01 & \textbf{63.96} & 62.83 & \textbf{64.32} \\ \hline
\textbf{QNLI} & Accuracy & 61.1 & \textbf{61.98} & 61.77 & \textbf{62.29} & 61.99 & \textbf{63.12} \\ \hline
\textbf{RTE} & Accuracy & 50.31 & \textbf{50.64} & 50.73 & \textbf{50.88} & 51.3 & \textbf{52.7} \\ \hline
\textbf{WNLI} & Accuracy & 50.23 & \textbf{50.78} & 51.03 & \textbf{51.28} & 51.65 & \textbf{52.84} \\ \hline
\multicolumn{2}{|c|}{\textbf{Average}} & 63.14 & \textbf{64.26} & 64.08 & \textbf{65.02} & 64.86 & \textbf{66.17} \\ \hline
\end{tabular}
\end{adjustbox}
\end{table*}

\subsection{Experimental Setup}
We provide detailed descriptions of experimental settings.
\subsubsection{Pre-training Settings} We pre-train both DA-MoE and Switch Transformer (ST) on the WikiText-103 dataset~\cite{b9}, which is a collection of 103 million tokens extracted from Wikipedia articles and is widely used for language modeling tasks due to its high-quality and well-curated text. As suggested in~\cite{b5}, we configure both DA-MoE and ST models with 8, 16, and 32 experts, 768-dimensional embeddings, 3072-dimensional hidden layer representations, and 12 attention heads.

\begin{figure}
    \centering
    \includegraphics[width=1\linewidth]{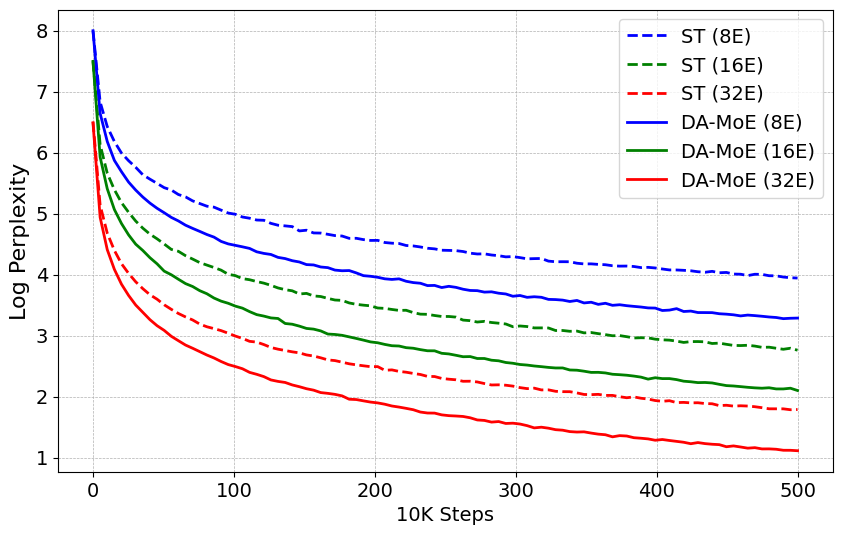}
    \caption{Training log perplexity scaling comparison between DA-MoE and ST (Switch Transformer) base models.}
    \label{fig:perplexity}
\end{figure}

\subsubsection{Fine-tuning Settings} We obtain the pre-trained ST-Base model weights, configured with 8, 16, and 32 experts, from~\cite{b21} to initialize both DA-MoE and ST models to be fined-tuned and evaluated on the GLUE benchmark.
The GLUE benchmark is a collection of various representative natural language processing tasks~\cite{b8}, including (1) single-sentence classification tasks: (i) SST-2~\cite{b12}, which is a binary classification task that classifies movie reviews as positive or negative, and (ii) CoLA (Corpus of Linguistic Acceptability)~\cite{b13}, which is a linguistic acceptability task, (2) similarity and paraphrase tasks: (i) MRPC (Microsoft Research Paraphrase Corpus)~\cite{b11}, which is a paraphrase detection task, and (ii) QQP (Quora Question Pairs), which is a duplication detection task, and four natural language inference tasks: (i) MNLI (Multi-Genre Natural Language Inference)~\cite{b14}, (ii) QNLI (Question Natural Language Inference)~\cite{b15}, (iii) RTE (Recognizing Textual Entailment)~\cite{b16,b17,b18,b19}, and WNLI (Winograd Schema Challenge)~\cite{b10}.
We perform three rounds of fine-tuning using different random seeds and report the average results on the validation sets.

\begin{table}[h]
\centering
\caption{Fine-tuning results on GLUE subtasks (Pre-trained on C4 dataset). The best result for each task is emphasized in bold.}
\label{table:fine-training-results-C4}
\begin{adjustbox}{max width=\textwidth}
\begin{tabular}{|c|c|c|c|}
\hline
\multirow{2}{*}{\textbf{GLUE Subtasks}} & \multirow{2}{*}{\textbf{Metrics}} & \multicolumn{2}{c|}{\textbf{Our Experiments}} \\ \cline{3-4} 
 &  & \textbf{Baseline} & \textbf{Our approach} \\ \hline \hline
\textbf{CoLA} & Accuracy & \textbf{70.41} & 68.78 \\ \hline
\textbf{SST-2} & Accuracy & 93.54 & \textbf{94.19} \\ \hline
\textbf{MRPC} & F1 & 82.33 & \textbf{83.82} \\ \hline
\textbf{QQP} & F1 & 88.13 & \textbf{89.57} \\ \hline
\textbf{MNLI} & Accuracy & 85.78 & \textbf{86.97} \\ \hline
\textbf{QNLI} & Accuracy & 89.05 & \textbf{90.06} \\ \hline
\textbf{RTE} & Accuracy & \textbf{58.21} & 56.62 \\ \hline
\textbf{WNLI} & Accuracy & 51.57 & \textbf{51.83} \\ \hline
\end{tabular}
\end{adjustbox}
\end{table} 

\begin{figure*}[!h]
    \centering
    \includegraphics[width=1\linewidth]{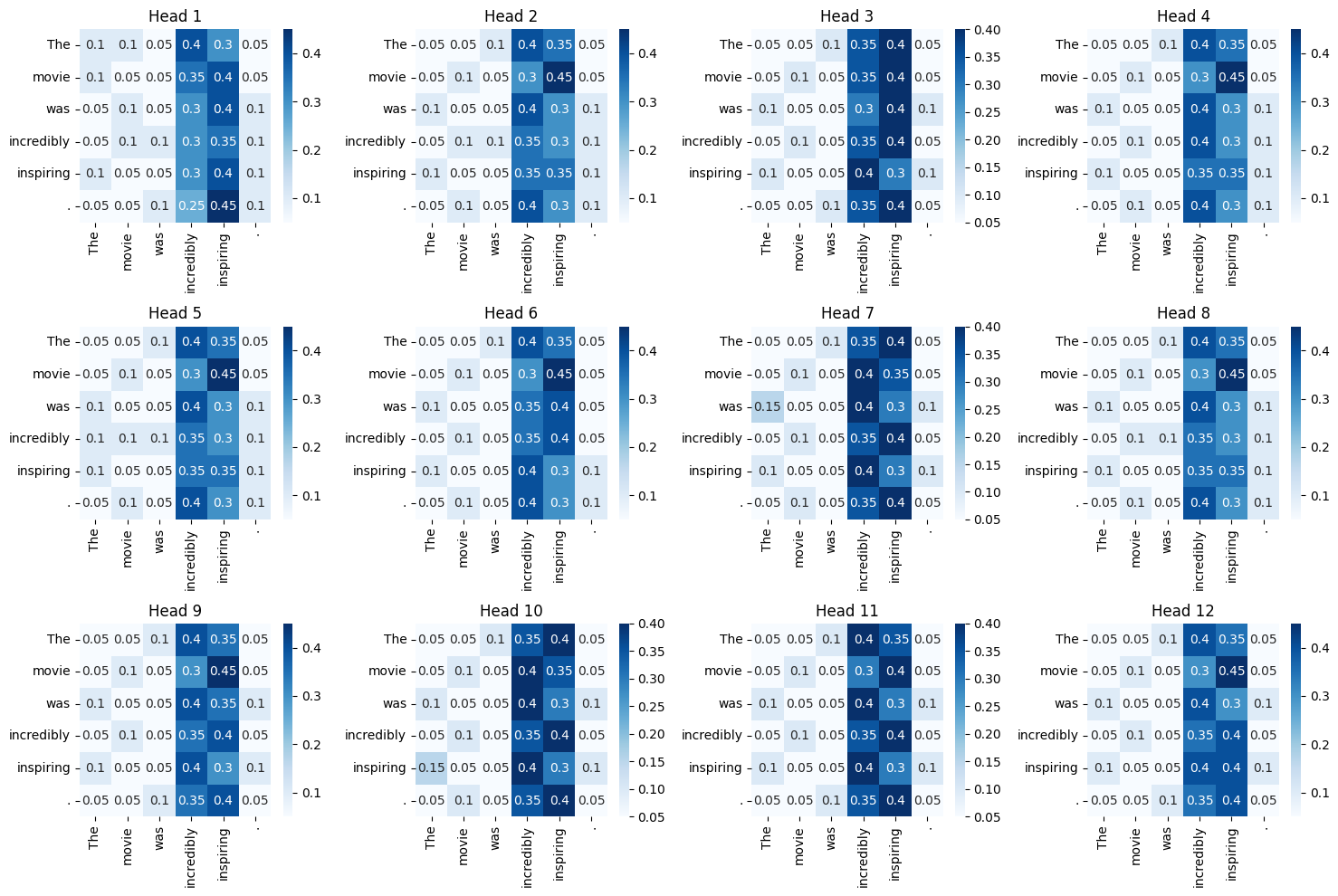}
    \caption{Attention weights for 12 heads in the last layer for sentiment analysis task for the sentence (``The movie was incredibly inspiring.")}
    \label{fig:attention-weights-sentiment-analysis}
    \vspace{-2mm}
\end{figure*}

\begin{figure*}[!h]
    \centering
    \includegraphics[width=1\linewidth]{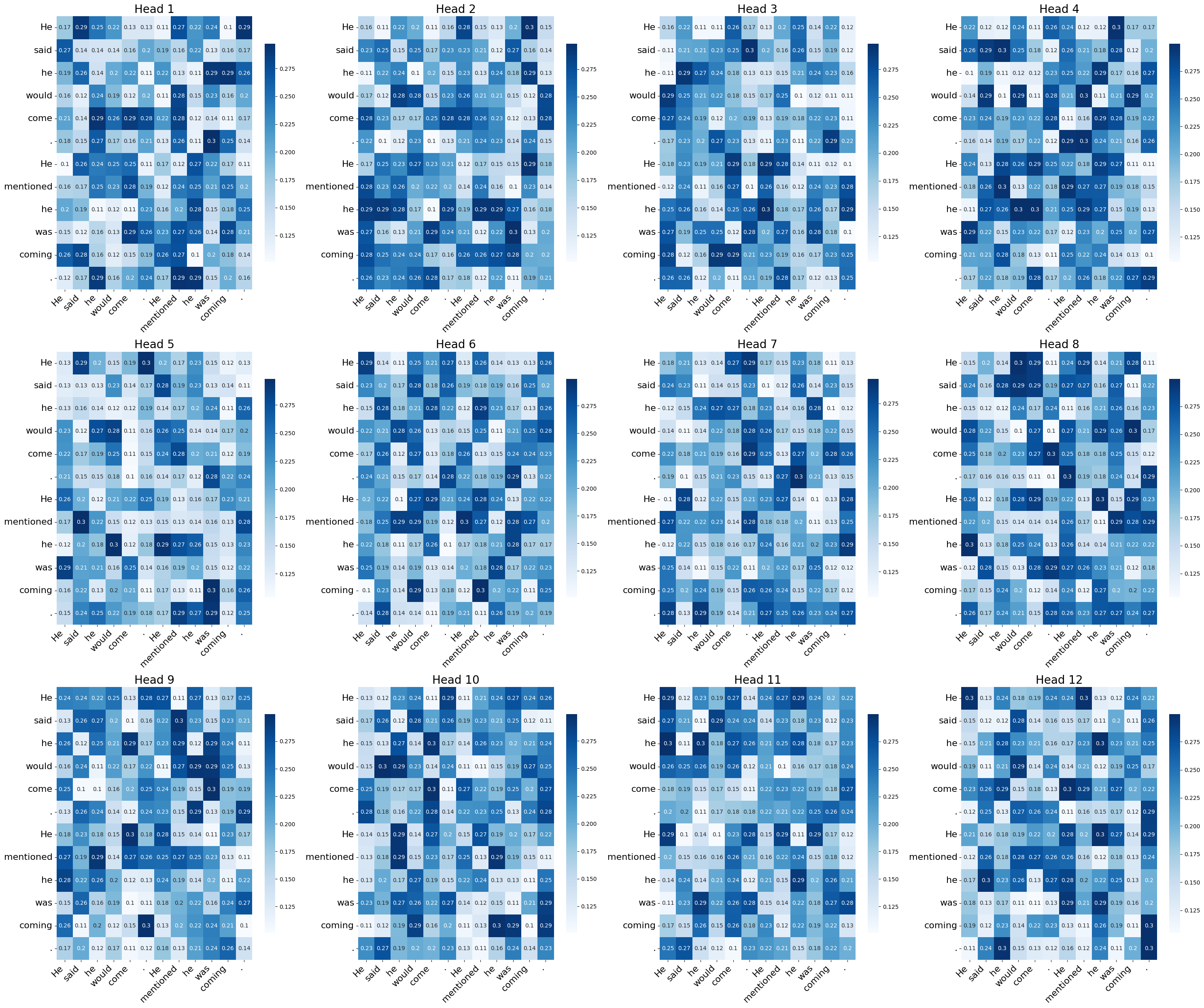}
    \caption{Attention weights for 12 heads in the last layer for paraphrasing task for the sentence (``He said he would come. He mentioned he was coming.")}
    \label{fig:attention-weights-paraphrasing}
    \vspace{-2mm}
\end{figure*}
\begin{figure*}[!h]
    \centering
    \begin{subfigure}{0.49\linewidth}
        \centering
        \includegraphics[width=\linewidth]{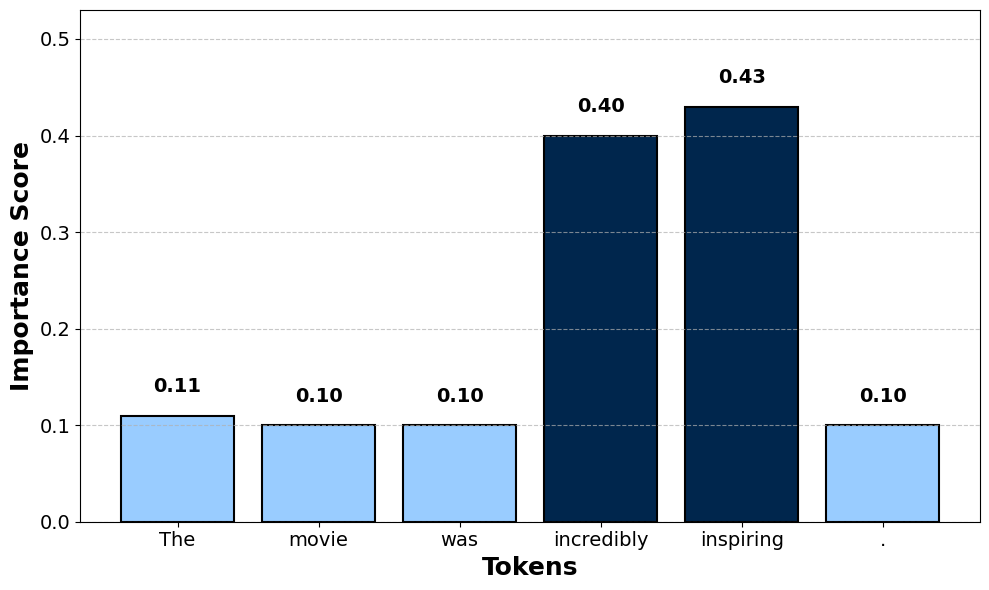}
        \caption{Token importance for sentiment analysis task for the sentence (``The movie was incredibly inspiring.")}
        \label{fig:token_importance_sentiment_analysis}
    \end{subfigure}
    \hfill
    \begin{subfigure}{0.49\linewidth}
        \centering
        \includegraphics[width=\linewidth]{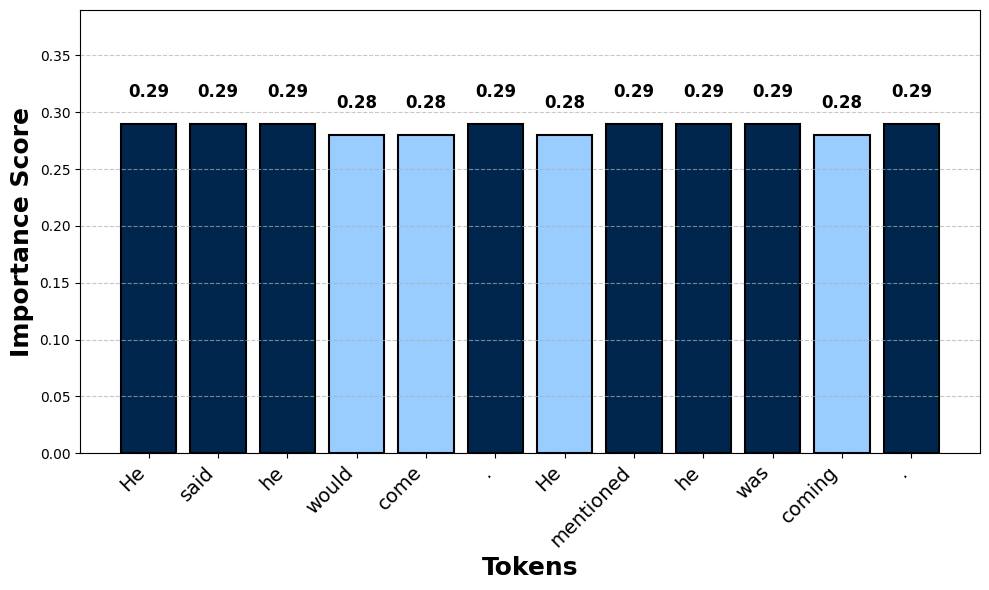}
        \caption{Token importance for paraphrasing task for the sentence (``He said he would come. He mentioned he was coming.")}
        \label{fig:token_importance_paraphrasing_task}
    \end{subfigure}
    \caption{Token importance across (a) sentiment analysis and (b) paraphrasing task examples.}
    \label{fig:token_importance_comparison}
\end{figure*}

\subsection{Pre-training Evaluation}
We first pre-train both DA-MoE and ST models with scaling the number of experts from 8 to 16 and 32 on the WikiText-103 dataset and fine-tuning these pre-trained models for the GLUE benchmark evaluation. Figure~\ref{fig:perplexity} presents the training log perplexity, while Table~\ref{table:pre-training-results-WikiText-103} presents the evaluation results using the GLUE benchmark. We highlight two interesting observations.
{\it First,} our DA-MoE approach significantly outperforms the baseline ST method in 7 out of a total 8 GLUE benchmark tasks except for MRPC, achieving accuracy/F1 score improvements of up to 8.58\% (the 2B model with 32 experts, 2B/32E, the last two columns). These experimental results demonstrate that our DA-MoE approach can significantly improve the performance of the baseline ST model across various NLP tasks. 
{\it Second,} the proposed DA-MoE approach demonstrates enhanced scalability. We evaluated the model performance of both DA-MoE and ST models by scaling the number of exerts from 8 to 16 and 32. The experimental results indicate that increasing the number of experts (i.e., expanding the model size) effectively reduces the log perplexity during training in Figure~\ref{fig:perplexity}, and significantly improves NLP task performance in Table~\ref{table:pre-training-results-WikiText-103}.
Compared to the baseline ST method, our DA-MoE approach consistently delivers enhanced performance across various NLP tasks.
{\it Third,} the average performance on the GLUE benchmark shows that our DA-MoE substantially outperforms the baseline ST models by over 0.94\%. This further demonstrates that the proposed dynamic expert allocation based on token importance effectively enhances model performance in handling various NLP tasks compared to the conventional routing mechanism with a fixed number of experts adopted in the baseline ST models.

\subsection{Fine-tuning Evaluation}
We then evaluate our DA-MoE approach through language model fine-tuning. We obtain the pre-trained ST model weights with 16 experts from the official ST model repository on Hugging Face~\cite{b21}, pre-trained on the C4 dataset~\cite{b20}. Table~\ref{table:fine-training-results-C4} presents the evaluation results on the GLUE benchmark after fine-tuning both our DA-MoE and the baseline ST models initialized with these pre-trained weights. 
Three interesting observations should be highlighted.
{\it First,} the proposed DA-MoE approach effectively outperforms the baseline ST method in 6 out of 8 subtasks in the GLUE benchmark except for CoLA and RTE. This shows that our DA-MoE approach can effectively enhance the language model performance during fine-tuning as well.
{\it Second,} fine-tuning language models pre-trained on a larger dataset (e.g., C4) can significantly improve NLP task performance. Compared to the pre-training results on the smaller dataset, WikiText-103, in Table~\ref{table:pre-training-results-WikiText-103}, fine-tuning pre-trained models on the C4 dataset (see Table~\ref{table:fine-training-results-C4}) can significantly enhance the NLP task accuracy/F1 score by up to 27.77\% (on QNLI by our DA-MoE).
{\it Third,} for both language model pre-training and fine-tuning, our DA-MoE substantially enhances the NLP task performance across the majority of subtasks in the GLUE benchmark. This further demonstrates that the proposed dynamic expert allocation effectively optimizes the conventional Mixture-of-Experts models with a fixed number of experts to more efficiently capture the latent semantic information from tokens of high importance.

\subsection{Token Importance Analysis}

In this section, we investigate how the proposed DA-MoE approach captures and utilizes the token importance in improving language model predictive performance. We identified two examples: (1) ``The movie was incredibly inspiring." for sentiment analysis from SST-2 and (2) the sentence pair ``He said he would come. He mentioned he was coming." for the paraphrasing task from MRPC. Figure~\ref{fig:attention-weights-sentiment-analysis} and Figure~\ref{fig:attention-weights-paraphrasing} illustrate the attention weights for each example sentence across all 12 attention heads from our DA-MoE model, pre-trained on the WikiText-103 dataset. Figure~\ref{fig:token_importance_sentiment_analysis} and Figure~\ref{fig:token_importance_paraphrasing_task} present the corresponding token importance scores for each word in the example sentences. 
We found three interesting observations. 

{\it First,} our DA-MoE model demonstrated a strong ability to identify and focus on sentiment-bearing tokens. For the sentence ``The movie was incredibly inspiring.", our DA-MoE model assigned high token importance scores to two tokens, i.e., ``incredibly" and ``inspiring". This demonstrates that DA-MoE effectively recognizes these words as key sentiment indicators (see Figure~\ref{fig:token_importance_sentiment_analysis}).
{\it Second,} for paraphrasing tasks, such as MRPC (Microsoft Research Paraphrase Corpus), where the model needs to determine if two sentences are semantically equivalent, we observed a more uniform distribution of token importance scores across input tokens. For the sentence pair, ``He said he would come. He mentioned he was coming." each of the twelve tokens received an importance score of approximately 0.3, with no single token deemed unimportant (see Figure~\ref{fig:token_importance_paraphrasing_task}). This suggests that the model attention should be evenly distributed across all input tokens.
{\it Third,} comparing Figure~\ref{fig:attention-weights-sentiment-analysis} with Figure~\ref{fig:token_importance_sentiment_analysis} and Figure~\ref{fig:attention-weights-paraphrasing} with Figure~\ref{fig:token_importance_paraphrasing_task}, we observe that the attention weights accurately reflect the relative token importance, which is consistent with findings in the literature~\cite{b1,b2,b37}. This further demonstrates that our token importance score calculation based on attention weights is effective in capturing semantically significant tokens, contributing to improved language model predictive performance.

\section{Conclusion} \label{section:conclusion}
This paper presents a novel dynamic expert allocation mechanism for Mixture-of-Experts (MoE) models. We made three original contributions.
{\it First,} we show that the existing MoE techniques with a fixed number of experts allocation may not efficiently adapt to tokens of varying importance, leading to inefficient computational resource usage and sub-optimal predictions.
{\it Second,} we propose a new router mechanism in MoE models that dynamically allocates experts to input tokens based on their importance, computed from the Transformer attention layer, which can effectively enhance overall efficiency and predictive performance.
{\it Third,} we perform extensive experiments on representative benchmark datasets for both pre-training and fine-tuning evaluations, which demonstrates that the proposed DA-MoE model effectively scales to multiple experts and significantly outperforms the state-of-the-art Transformer MoE on the GLUE benchmark.
The proposed DA-MoE is a generic approach that can be applied to various MoE architectures. Our future work will explore the integration of our DA-MoE into other types of MoE models and perform further experimental analysis on other large-scale datasets.

\bibliographystyle{IEEEtran}
\bibliography{references}

\end{document}